\documentclass{vgtc}                          

\ifpdf
  \pdfoutput=1\relax                   
  \usepackage{graphicx}                
  \DeclareGraphicsExtensions{.pdf,.png,.jpg,.jpeg} 
\else
  \usepackage{graphicx}                
  \DeclareGraphicsExtensions{.eps}     
\fi%

\graphicspath{{figures/}{pictures/}{images/}{./}} 

\usepackage{microtype}                 
\PassOptionsToPackage{warn}{textcomp}  
\usepackage{textcomp}                  
\usepackage{mathptmx}                  
\usepackage{times}                     
\usepackage{cite}                      
\usepackage{tabu}                      
\usepackage{booktabs}                  

\usepackage{caption}
\usepackage{bm}
\usepackage{amssymb}
\usepackage{mathtools}
\usepackage{multirow}
\usepackage{makecell}
\usepackage{subcaption}
\usepackage[normalem]{ulem}

\usepackage{color}

\newcommand{\reffig}[1]{Fig.~\ref{#1}}
\renewcommand{\refeq}[1]{Eq.~(\ref{#1})}
\newcommand{\reftab}[1]{Table~\ref{#1}}
\newcommand{\refsec}[1]{Sec.~\ref{#1}}

\usepackage[acronym,shortcuts,acronymlists={hidden}]{glossaries}
\newacronym{AR}{AR}{Augmented Reality}
\newacronym{ARVR}{AR/VR}{Augmented and Virtual Reality}
\newacronym{ATE}{ATE}{absolute trajectory error}
\newacronym{AWS}{AWS}{Amazon Web Services}
\newacronym{DoF}{DoF}{degrees of freedom}
\newacronym{GBA}{GBA}{Global Bundle Adjustment}
\newacronym{IMU}{IMU}{Inertial Measurement Unit}
\newacronym{KF}{KF}{Keyframe}
\newacronym{LM}{LM}{Landmark}
\newacronym{p2p}{p2p}{peer-to-peer}
\newacronym{PGO}{PGO}{Pose-Graph Optimization}
\newacronym{RBCD}{RBCD}{Riemannian Block-Coordinate Descent}
\newacronym{RMSE}{RMSE}{Root Mean Squared Error}
\newacronym{SfM}{SfM}{Structure from Motion}
\newacronym{SLAM}{SLAM}{Simultaneous Localization And Mapping}
\newacronym{TSDF}{TSDF}{Truncated Signed Distance Field}
\newacronym{UAV}{UAV}{Unmanned Aerial Vehicle}
\newacronym{VIO}{VIO}{Visual-Inertial Odometry}

\onlineid{0}

\vgtccategory{Research}

\vgtcinsertpkg

\title{COVINS: Visual-Inertial SLAM for Centralized Collaboration}

\author{Patrik Schmuck
\and Thomas Ziegler
\and Marco Karrer
\and Jonathan Perraudin
\and Margarita Chli
\thanks{
This work was supported by SNSF (Agreement no. PP00P2 157585) and NCCR Robotics.}
}
\affiliation{\scriptsize Vision for Robotics Lab, ETH Zurich, Switzerland}

\abstract{
Collaborative SLAM enables a group of agents to simultaneously co-localize and jointly map an environment, thus paving the way to wide-ranging applications of multi-robot perception and multi-user \acs{AR} experiences by eliminating the need for external infrastructure or pre-built maps.
This article presents COVINS, a novel collaborative SLAM system, that enables multi-agent, scalable SLAM in large environments and for large teams of more than 10 agents.
The paradigm here is that each agent runs visual-inertial odomety independently onboard in order to ensure its autonomy, while sharing map information with the COVINS server back-end running on a powerful local PC or a remote cloud server.
The server back-end establishes an accurate collaborative global estimate from the contributed data, refining the joint estimate by means of place recognition, global optimization and removal of redundant data, in order to ensure an accurate, but also efficient SLAM process.
A thorough evaluation of COVINS reveals increased accuracy of the collaborative SLAM estimates, as well as efficiency in both removing redundant information and reducing the coordination overhead, and demonstrates successful operation in a large-scale mission with 12 agents jointly performing SLAM. 
} 

\keywords{
Collaborative SLAM, Computer Vision, Multi-Agent Systems, Augmented and Mixed Reality, Large-scale SLAM
\vspace{-3pt}
}

\CCScatlist{
  \CCScatTwelve{Computing methodologies}{Artificial intelligence}{Computer vision}{Tracking and Reconstruction}
\vspace{-3pt}
}

\begin{document}


\firstsection{Introduction}

\maketitle
	With \ac{SLAM} having reached significant maturity and robustness, not only it has started being employed in more costumer products, but also more complex, multi-agent applications have been gaining traction in the research community.
	Sharing information amongst participants and dividing up tasks between multiple agents promises to boost robustness, efficiency and accuracy of a robotic mission in various scenarios.
	Beyond robotics, recently emerging technologies, such as \ac{ARVR} depend on collaborative perception systems in order to create shared experiences for multiple users.
	Particularly, this vision of multi-user shared \acs{AR} currently experiences high attention, with companies such as Microsoft, Magic Leap or Nreal actively researching and developing such solutions.
	Similar to single-agent scenarios, a large body of the collaborative perception literature focuses on either mapping 	\cite{guo2018:TRO:resource}	\textit{or} localization \cite{jung2020:ICRA:decentralized}	from multiple agents.
    However, it is only when addressing both challenges simultaneously, that collaborative \ac{SLAM} can happen, and eventually enable the full spectrum of possibilities a multi-agent system has to offer.
	Seamless multi-user \acs{AR} experience requires collaborative SLAM for maximum flexibility, avoiding tedious pre-mapping or putting up external tracking infrastructure, as well as the efficient deployment of robotic teams in search-and-rescue situations, where generating an initial map would significantly decrease response time. \\
	\begin{figure}[t]
		\centering
        \includegraphics[width=0.99\columnwidth]{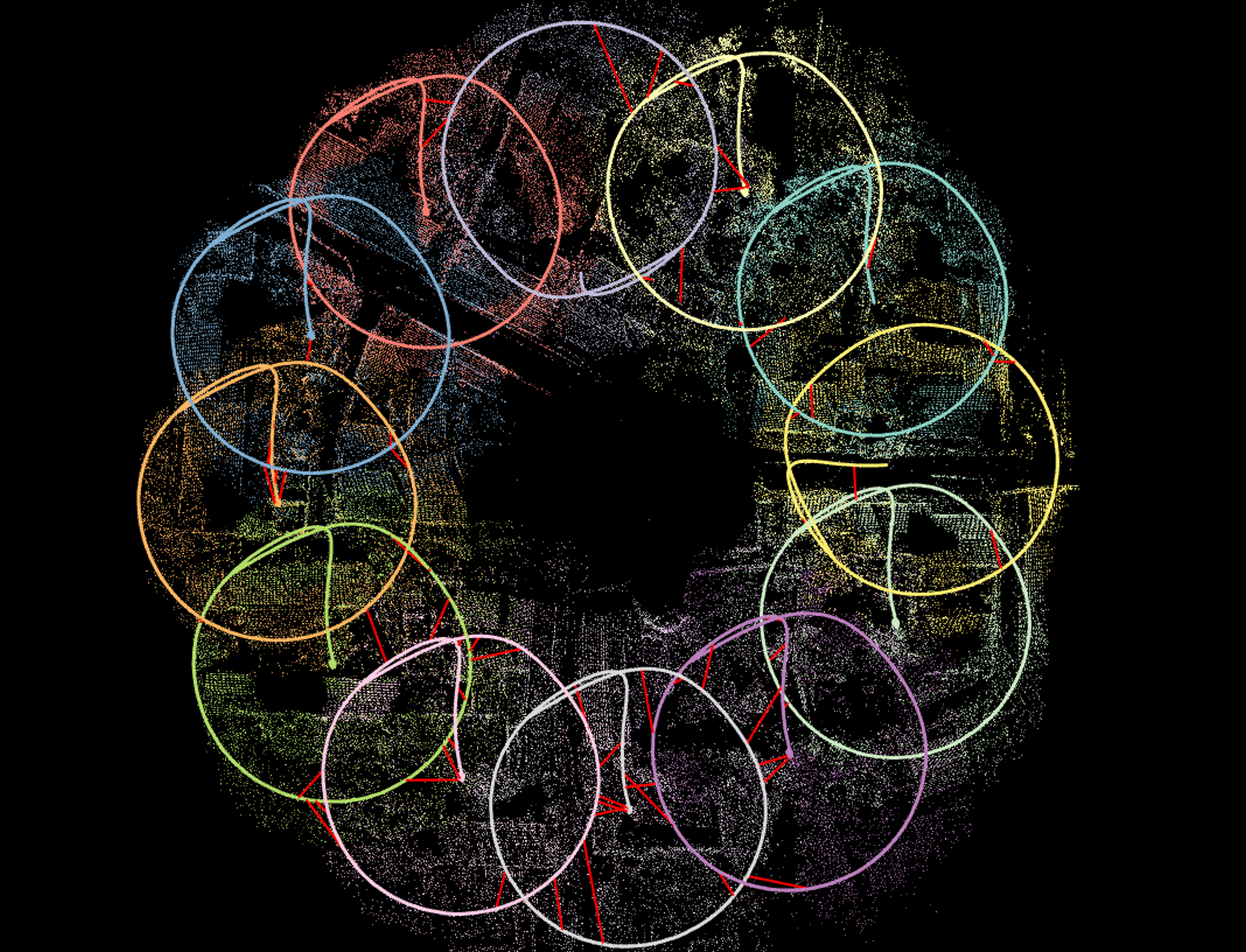}
        \vspace{-5pt}
		\caption{
			Collaborative \ac{SLAM} estimate from 12 drones flying over a village scenery 
			covering approx. $500 m^2$ with $1750 m$ total trajectory length. Red lines indicate inter-trajectory constraints.
		}
		\label{fig:intro:intropic}
	    \vspace{-20pt}
	\end{figure}
	At the same time, multi-agent \ac{SLAM} poses significant challenges, such as accurate co-localization,  ensuring consistency with multiple agents simultaneously contributing information, and managing scalability with regard to the large amount of contributed data.
	Several works have shown good progress tackling one ore more of these challenges over the last few years \cite{chang2020:ARXIV:kimera,karrer2018cvi,lajoie2020:RAL:door,schmuck2018:JFR:ccm,schmuck2019:3DV:redundancy}.
	However, collaborative \ac{SLAM} is a relatively young research field, albeit a very promising one.
	In this spirit, this work presents COVINS, a collaborative Visual-Inertial (VI) SLAM system, enabling a group of agents, each equipped with a VI sensor suite, to establish collaborative scene understanding online during a mission, through co-localization and joint creation of a global map of the environment.
	Together with the ability to share data through the server, this provides the basis to deploy coordinated multi-robot missions and shared \acs{AR} experiences.
	\\
	The core of COVINS constitutes a comprehensive revision of a well-established architecture approach for multi-agent \ac{SLAM} \cite{karrer2018cvi} as well as of the individual system modules,
	distilling the best aspects of the highest performing modules from the state of the art, revisiting the key ideas behind them, as well as their interaction.
	This translates into increased versatility and efficiency of the framework, and enables scalability to larger groups of agents.
	This framework is shown to achieve improved accuracy, and allows to demonstrate collaborate \ac{SLAM} in a scenario with up to 12 agents contributing simultaneously to the system (\reffig{fig:intro:intropic}), which to the best of our knowledge is the most populous team demonstrating to perform collaborative VI \ac{SLAM} to date.
	%
    The implementation of COVINS will be released for public use with the final version if this article.

\section{Related Work}
	%
	%
	The literature in multi-agent perception systems distinguishes between centralized and distributed architectures.
	One of the first works to tackle collaborative \ac{SLAM} in a fully decentralized manner was DDF-SAM \cite{cunningham2013ddf}, evaluating robotic collaboration in a simulated setup using visual, inertial and GPS data. 
	Choudhary et al. \cite{choudhary2017:IJRR:distributed} show a decentralized \ac{SLAM} system with pre-trained objects enabling co-localization of the participating robots.
	Cieslewski et al. \cite{cieslewski2018:ICRA:data} combine this optimization approach with an efficient and scalable distributed solution for place recognition.
	Most recently, Lajoie et al. \cite{lajoie2020:RAL:door} and Chang et al. \cite{chang2020:ARXIV:kimera} proposed systems for distributed \ac{SLAM} both using a distributed \ac{PGO} scheme, demonstrating superior performance compared to the Gauss-Seidel approach of \cite{choudhary2017:IJRR:distributed}.
	%
	%
	While enabling a wide range of applications, and good scalability to large numbers of agents, guaranteeing data consistency and avoiding information double-counting are the biggest challenges for this architecture, whereas centralized systems have a more straightforward management of information, and usually exhibit a significantly higher accuracy \cite{campos2021:TRO:orb3,karrer2018cvi,qin2018:TRO:vins} than state-of-the-art distributed \ac{SLAM} approaches, such as \cite{chang2020:ARXIV:kimera,lajoie2020:RAL:door}.
	\\
	%
	Zou and Tan have introduced CoSLAM \cite{zou2013coslam}, a powerful vision-only collaborative \ac{SLAM} system, grouping cameras with scene overlap in order to handle dynamic environments.
	Forster at al. \cite{forster2013collaborative} demonstrated collaboration of up to three \acp{UAV} by extending a \ac{SfM} pipeline to collaborative \ac{SLAM}.
	With C${}^2$TAM \cite{riazuelo2014c}, Riazuelo et al. proposed a multi-agent system performing only position tracking onboard each agent, while all mapping tasks are offloaded to the server,
	enabling agents to cope with very limited computational resources, however, thereby also heavily restricting each agent's autonomy.
	CCM-SLAM \cite{schmuck2018:JFR:ccm} proposed to efficiently make use of the server by offloading computationally expensive tasks, while still ensuring each agent's autonomy at low computational resource requirements by running a visual odometry system onboard.
	CVI-SLAM \cite{karrer2018cvi} extends the approach of \cite{schmuck2018:JFR:ccm} to a visual-inertial setup, enabling higher accuracy as well as metric scale estimation and gravity alignment of the collaborative \ac{SLAM} estimate, demonstrated with real data from up to four agents.
	However, while it was the first full VI collaborative \ac{SLAM} system with two-way communication, CVI-SLAM also has practical limitations, such as limited flexibility, e.g. in terms of interfacing with custom \ac{VIO} front-ends, and in terms of scalability to larger teams.
	\\
	%
	The ability to leverage \ac{VIO} to enable \acs{AR} experiences with mobile devices has been shown by multiple works over the last years, such as \cite{li2017:ISMAR:monocular, qin2018:TRO:vins}.
	Just recently, Platinsky et al. \cite{platinsky2020:ISMAR:collaborative} have demonstrated a pipeline supporting city-scale shared augmented reality experiences on mobile devices. 
	However, their approach relies on time-consuming and costly preparatory work, comprising extensive data collection using a car-based platform and offline map generation.
	On the other hand, the concept of \textit{spatial anchors} can enable ad hoc multi-user \acs{AR}, through co-localization	with respect to the same anchor.
	However, this requires at least coarse prior knowledge of the location of the users, for example knowing all devices are in the same room, and does not directly give individual agents the ability to re-use maps created by other agents.
	Collaborative \ac{SLAM} can bridge the gap between these two approaches, enabling shared \acs{AR} experience in larger environments, such as entire factory halls or department stores in an ad hoc fashion, only using the \acs{AR} devices' built-in sensors and without pre-mapping.
	\\
	%
	Multi-agent global collaborative estimates similar to those from collaborative \ac{SLAM} can also be achieved by recent SLAM systems with multi-session capabilities, such as \cite{campos2021:TRO:orb3,qin2018:TRO:vins}.
	The ability to re-use \ac{SLAM} maps created in a previous run enables multi-session SLAM systems to achieve impressive levels of robustness \cite{qin2018:TRO:vins} and accuracy \cite{campos2021:TRO:orb3}, outperforming the single-session case.
	However, the circumstance that only one agent at a time can be active in multi-session \ac{SLAM} heavily restricts the level of collaboration amongst agents, and the situation that all parts of these multi-session \ac{SLAM} systems run onboard the same computing unit prevents to offload information and computation load to a powerful server.
	\\
	%
	COVINS extends the well-established architecture for collaborative SLAM deployed in \cite{karrer2018cvi} towards a more flexible and efficient setup.
	Agents can connect to the system on-the-fly, the number of agents does not need to be known {\it{a priori}}, and a generic communication interface allows to interface the server back-end with different \ac{VIO} systems.
	More efficient map management and optimization schemes and a state-of-the-art redundancy detection scheme translate into improved accuracy and better scalability, allowing to demonstrate collaborative SLAM with 12 agents, while to the best of our knowledge, other recent collaborative \cite{forster2013collaborative,karrer2018cvi,schmuck2018:JFR:ccm} and multi-session \cite{campos2021:TRO:orb3,qin2018:TRO:vins} \ac{SLAM} systems with comparable accuracy level are tested with no more than five agents so far.
\section{Preliminaries}
    \subsection{Notation, IMU Model and System States} \label{sec:prelim:notation_and_state}
        In this work, we adopt the notation from \cite{karrer2018cvi} for mathematical notation.
        Small (e.g. $\mathbf{a}$) and large (e.g. $\mathbf{A}$) bold letters denote vectors and matrices, respectively. 
        Coordinate frames are denoted as plain capital letters (e.g. $A$).
        For a vector $\mathbf{x}$ expressed in $A$, the notation $\prescript{}{A}{\mathbf{x}}$ is used.
        A rigid body transformation from frame $B$ to $A$ is denoted as $\mathbf{T}_{AB}$, with $\mathbf{t}$ and $\mathbf{R}$ denoting the translational and rotational part, respectively.
        Throughout this work we denote the world frame as $W$, the \ac{IMU} body frame as $S$ and the camera frame as $C$.
        \\
        In order to incorporate \ac{IMU} information into COVINS, we model the \ac{IMU} using a standard model, assuming additive Gaussian noise and unknown, time varying sensor bias (cf. \cite{covins:supp}).
        To account for this \ac{IMU} model, the system state $\pmb{\Theta}$ includes besides the \ac{KF} poses and \ac{LM} positions also the linear velocities $_{W}\mathbf{v}$ as well as the bias variables for each \ac{KF} $k$:
        \vspace{-5pt}
        \begin{equation}\label{eq:system_state}
        \pmb{\Theta} \coloneqq \{ \underbrace{\mathbf{R}_{WS}^{k}, \mathbf{t}_{WS}^{k}, _{W}\mathbf{v}^{k}, \mathbf{b}^{k}}_{\text{KF}_{k}}, \prescript{}{W}{\mathbf{l}}^{i} \}, \forall k \in \mathcal{V}, \forall i \in \mathcal{L} ~,
        \vspace{-5pt}
        \end{equation}
        where the sets $\mathcal{V}$ and $\mathcal{L}$ denote the set of all \acp{KF} and \acp{LM}, respectively.
        In the following, whenever the context allows for it, we use $\theta_{j}$ to denote an individual state variable.
\section{Methodology}
	\subsection{System Overview}
		\begin{figure}[t]
			\centering
			\includegraphics[scale=0.45]{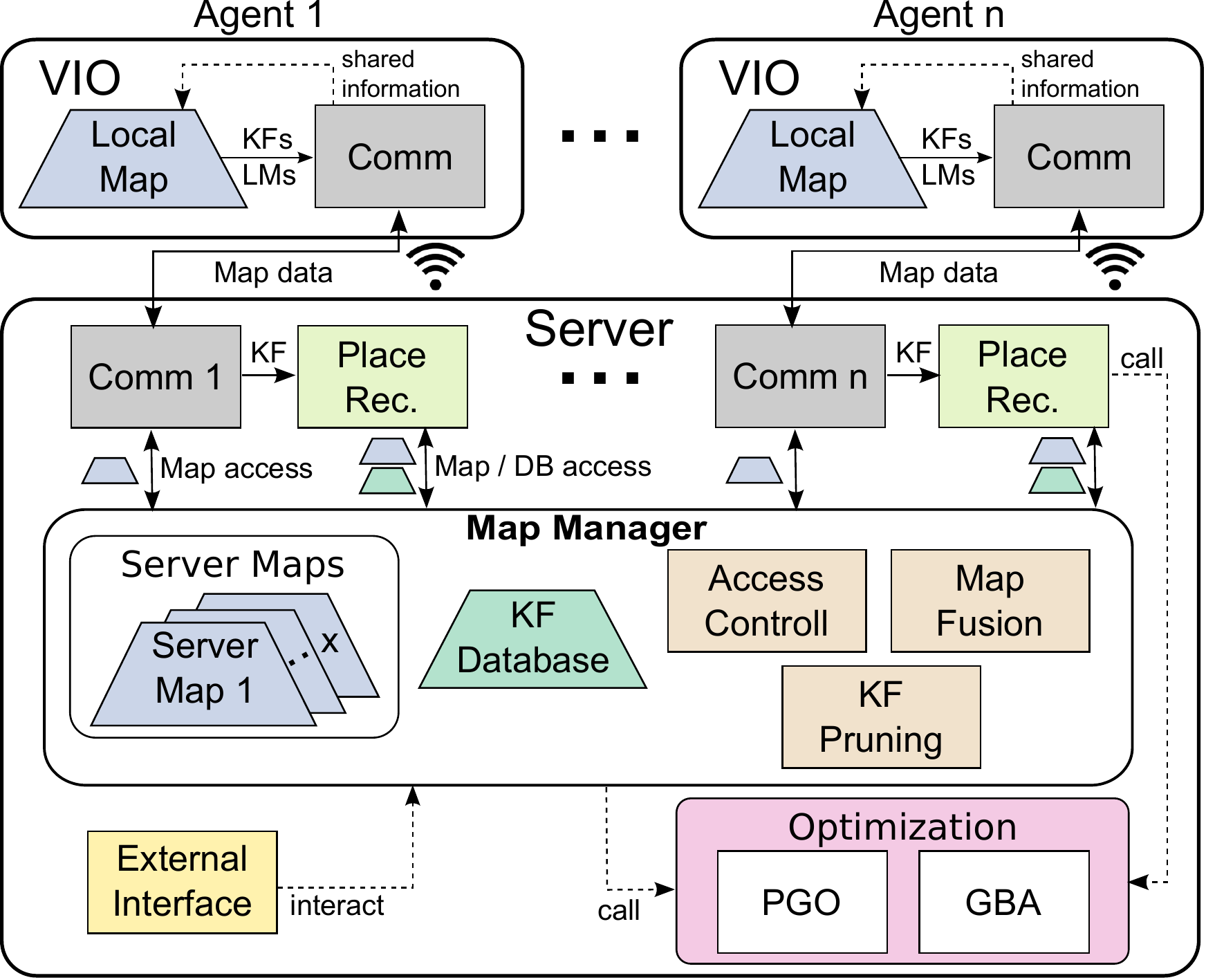}
			\vspace{-5pt}
			\caption{
				Overview of the COVINS system architecture. 
			}
			\label{fix:meth:sysarch}
			\vspace{-15pt}
		\end{figure}
		The system architecture of COVINS is illustrated in \reffig{fix:meth:sysarch}. 
		Running a \ac{VIO} front-end maintaining a local map of limited size onboard each agent ensures basic autonomy of the individual agent.
		At the same time, global map maintenance and computation-heavy processes are transferred to the more powerful server.
		This underlying architectural principle was first introduced in \cite{schmuck2017multiuav}, which is extended to a more flexible and efficient handling and maintenance of collaborative map data on the server in this work.
		On both the agents' and server's sides, we deploy a communication module for data exchange.
		The communication module establishes \ac{p2p} connection, allowing the server to run on a locally deployed computer as well as on a remote cloud service, as demonstrated in \refsec{sec:exp}, furthermore removing the previous ROS dependency of the communication module in \cite{karrer2018cvi}.
		COVINS implements and exports a generic communication interface, providing the freedom to interface it with any custom-built keyframe-based \ac{VIO} system, in order to enable collaboration amongst multiple agents. 
		The core of the server modules forms a map manager, which controls access to the global map data present in the system. 
		It maintains this map data in one or multiple maps, as well as a \ac{KF} database for efficient place recognition. Moreover, it provides algorithms to merge maps once overlap is detected and the functionality to remove redundant \acp{KF}, altogether facilitating data routing compared to \cite{karrer2018cvi}. 
		Place recognition modules process all incoming \acp{KF} from the agents to detect visual overlap between re-visited parts of the environment.
		As opposed to \cite{karrer2018cvi}, COVINS does not distinguish between different place recognition modules for either loop closure or map fusion, so a single place recognition query for a \ac{KF} triggers both events, reducing workload and system complexity.
		The server, furthermore, provides optimization routines, namely \acf{PGO} and \acf{GBA}.
		In contrast to \cite{karrer2018cvi}, COVINS implements an optimization strategy regularly performing \ac{PGO}, while executing \ac{GBA} to further refine maps less frequently, in order to better balance restricted map access due to ongoing optimization with desired high map accuracy.
		In addition, the server provides an external interface, allowing a user to interact with the system.
	\subsection{Map Structure} \label{sec:meth:map}
		The map structure used by the server back-end of COVINS (termed server map) maintains the data of the collaborative estimation process.
		A server map $\mathcal{M}_{x}$ is a \ac{SLAM} graph, holding a set $\mathcal{V}_x$ of \acp{KF} and a set $\mathcal{L}_x$ of \acp{LM} as vertices, and edges induced either between two \acp{KF} through \ac{IMU} constraints or between a \ac{KF} and a \ac{LM} as a landmark observation.
		While multiple agents can contribute to one server map, multiple server maps exist simultaneously until all participating agents are co-localized.
		Together with the state definition from \refsec{sec:prelim:notation_and_state}, this underling \ac{SLAM} estimation problem induces a factor graph \cite{covins:supp}, which forms the basis of the \ac{GBA} scheme explained in \refsec{sec:meth:maprefine}.
		The shared \ac{LM} observations create dependencies between \acp{KF} from multiple agents, while \ac{IMU} factors are only inserted between consecutive \acp{KF} created by the same agent.
		In an \acs{AR} use-case, the map would furthermore store \acs{AR} content created by users contributing to this map.
	\subsection{Error Residuals Formulation } \label{sec:meth:residuals}
        By formulating a set of residuals, the optimization of state variables occurring in \ac{KF}-based VI SLAM can be expressed as a weighted nonlinear least-squares problem.
        Each such residual $\mathbf{e}_{i}$ expresses the difference between the expected measurement based on the current state of the system and the actual measurement $\mathbf{z}_{i}$:
        \vspace{-5pt}
        \begin{equation}\label{eq:residual_form}
            \mathbf{e}_{i} \coloneqq \mathbf{z}_{i} - h_{i}(\mathcal{A}_{i}) ~,
        \vspace{-5pt}
        \end{equation}
        where $\mathcal{A}_{i}$ is the set of state variables $\theta_{j}$ relevant for measurement $\mathbf{z}_{i}$, and $h_{i}(\cdot)$ is the measurement function, predicting the measurement according to these state variables in $\mathcal{A}_{i}$.
        By collecting all occurring residual terms, the objective of the optimization can be expressed as:
        \vspace{-5pt}
        \begin{equation}
            \pmb{\Theta}^{*} = \underset{\pmb{\Theta}}{\mathrm{arg min}} \left\{ \sum\limits_{i} \lVert \mathbf{z}_{i} - h_{i}(\mathcal{A}_{i}) \rVert_{\mathbf{W}_{i}}^{2} \right\} ~,
        \vspace{-5pt}
        \end{equation}
        where $\lVert \mathbf{x} \rVert_{W}^{2} = \mathbf{x}^{T} \mathbf{W} \mathbf{x}$ denotes the squared Mahalanobis distance with the information matrix $\mathbf{W}$.
        Within our system, we essentially use three different types of residuals:
        reprojection residuals $\mathbf{e}_{r}$, relative pose residuals $\mathbf{e}_{\Delta \mathbf{T}}$, and IMU pre-integration residuals $\mathbf{e}_{\text{IMU}}$.
        A detailed description of the individual residuals can be found in \cite{covins:supp}.
	\subsection{Visual-Inertial Odometry Front-End}
		COVINS is able to generate accurate collaborative global estimates from map data contributed by a keyframe-based \ac{VIO} system (also referred to as \ac{VIO} front-end).
		To enable the sharing of map information between this \ac{VIO} front-end running onboard the agent, and the server, this framework provides a communication interface that enables the combination of the server back-end with any indirect \ac{VIO} system (i.e. using feature-based landmark correspondences, required for the reprojection residuals) as explained in \refsec{sec:meth:comm}.
		For handling the inertial data in \ac{GBA}, we use the estimates of the metric scale as well as the \ac{IMU} biases and the velocities of the \ac{VIO} as an initialization point.
		In order to evaluate the performance of COVINS, in the experiments conducted for this paper, we employ the \ac{VIO} front-end of ORB-SLAM3 \cite{campos2021:TRO:orb3}, as a nominal open-source option.
	\subsection{Communication} \label{sec:meth:comm}
		The communication module is based on socket programming using the TCP protocol, and the header-only library \textit{cereal}\cite{code:cereal} for the serialization of messages.
		This allows to deploy the server on a local computational unit as well as on remote cloud servers
		The communication module on the server side listens to a pre-defined port for incoming connection requests by the agents, allowing them to join dynamically during the mission without any prior specification of the number of participating agents.
		A generic communication interface for usage on the agent side is exported by COVINS as a shared library, enabling an existing \ac{VIO} system to share map data with the server back-end using pre-defined \ac{KF} and \ac{LM} messages.
	\subsubsection{Agent-to-Server Communication}
		For sharing map data from the agent to the server, COVINS adopts the efficient message passing scheme from \cite{schmuck2018:JFR:ccm}, which accounts for static parts of \acs{KF} and \acp{LM}, such as extracted 2D feature keypoints and related descriptors, and ensures this information is not repeatedly sent, in order to reduce the required network bandwidth.
		The communication scheme distinguishes between so-called `full' messages, comprising all relevant information for a \ac{KF} and \ac{LM}, including also static measurements (e.g. 2D keypoints), and significantly smaller update messages, where only changes in the state (e.g. modified \ac{KF} pose) are transmitted.
		All map data to be shared with the server is accumulated over a short time window, and communicated to the server batch-wise at a fixed frequency.
		The communication counterpart on the server side integrates the transmitted map information into the collaborative \ac{SLAM} estimate.
	\subsubsection{Server-to-Agent Communication (Map Re-Use)}
		The communication interface of COVINS supports two-way communication between the agents and the server, in this article applied to estimate the drift of an agent's \ac{VIO}.
		On the server, drift can be accounted for on a global scope through loop closure and subsequent optimization-based map refinement.
		In order to enable the agent to also account for this drift, we regularly share the server's estimated pose of the most recently created \ac{KF} of an agent with this agent.
		Comparing this drift-corrected pose estimate from the server with the estimated pose of the \ac{KF} in the local map allows to estimate a local odometry transformation $\text{T}_{odom}$ onboard the agent, quantifying the drift in the current pose estimate.
		With this scheme, the map of the local \ac{VIO} is not modified, leaving the smoothness of the \ac{VIO} unaffected, which is of substantial importance, for example when using the pose estimate in a feedback system for controlling a robot.
	\subsection{Multi-Map Management} \label{sec:meth:mapman}
		The map manager maintains the data contributed by all agents in one or more server maps, as described in \refsec{sec:meth:map}. 
		A new map is initialized for every agent that enters the system.
		As soon as place recognition detects overlap between two distinct maps, the map fusion routine of the map manager is triggered.
        Furthermore, the map manager holds and maintains the \ac{KF} database necessary for efficient place recognition.
		%
		%
		Besides providing routines for map fusion and graph compression through removal of redundant \acp{KF} (\refsec{sec:meth:maprefine}), the map manager is in charge of controlling access to the server maps and the \ac{KF} database, in order to ensure global consistency.
		Storing all maps at a central point in the system with individual modules requesting access to either read from or also modify a specific map	facilitates to coordinate map access from different system modules in order to keep maps consistent, e.g. when multiple agents contribute to a single server map, or to restrict map access in order to perform map fusion or optimization.
	\subsection{Place Recognition, Loop Closure \& Map Fusion} \label{sec:meth:placerec}
		To detect repeatedly visited locations with high precision, we employ a standard multi-stage place recognition pipeline which we briefly summarize here.
		For a query \ac{KF} $\textit{KF}_q$, a bag-of-words approach \cite{galvez2012bags} is employed to select a set $\mathcal{C}$ of potential matching candidates from all \acp{KF} in the system.
		After establishing feature correspondences between $\textit{KF}_q$ and all \acp{KF} in $\mathcal{C}$, a 3D-2D RANSAC scheme followed by a refinement step minimizing the reprojection error is applied to find a relative transformation $\bm{T}_{cq}$ between $\textit{KF}_q$ and a potential match $\textit{KF}_c \in \mathcal{C}$.
		Finally, $\bm{T}_{cq}$ is used to find additional \ac{LM} connections between $\textit{KF}_q$ and $\textit{KF}_c$
		A place recognition match is accepted, if for a $\textit{KF}_c$ throughout all stages, enough inliers are found.
		In the case that $\textit{KF}_q$ and  $\textit{KF}_c$ are part of the same server map, we perform loop closure, carrying out a \ac{PGO} in order to optimize the poses of the \acp{KF} in the map, 
		improving accuracy and reducing drift in the estimate.
		In the case that $\textit{KF}_q$ and  $\textit{KF}_c$ reside in different server maps, the map fusion routine of the map manager is triggered, aligning the map $\mathcal{M}_{q}$ of the query KF and the the map $\mathcal{M}_{c}$ of the candidate KF using $\bm{T}_{cq}$, finally replacing both maps with one new server map $\mathcal{M}_{cq}$ containing all \acp{KF} from $\mathcal{M}_{q}$ and $\mathcal{M}_{c}$.
        This involves also the fusion of duplicate \acp{LM}.
		In the process, potential \acs{AR} content in $\mathcal{M}_{q}$ would be transformed into the coordinate frame of  $\mathcal{M}_{cq}$, and combined together with the \acs{AR} content contained in $\mathcal{M}_{c}$.
		This also entails that after map fusion, AR content of $\mathcal{M}_{c}$ is now available to all users previously associated to $\mathcal{M}_{q}$, and vice versa.
	\subsection{Map Refinement} \label{sec:meth:maprefine}
		\subsubsection{Pose-Graph Optimization} \label{sec:meth:opt:pgo}
			\ac{PGO}\footnote{All optimization schemes of COVINS use the Ceres solver}
			is applied to a map when a new loop constraint between two \acp{KF} is added to this map after successful loop closure detection.
			We use the following objective function for \ac{PGO}, optimizing the pose of all KFs of the server map:
			\vspace{-5pt}
			\begin{equation} \label{eq:pose_graph_objective}
			    J_{PGO}(\pmb{\Theta}) \coloneqq \sum\limits_{i \in \mathcal{V}} \sum\limits_{j \in \mathcal{V}} x(i, j) \cdot \lVert \mathbf{e}_{\Delta\mathbf{T}}^{i, j} \rVert_{\mathbf{W}_{\Delta \mathbf{T}}}^{2} ~,
		    \vspace{-5pt}
			\end{equation}
			where $x(i, j)$ is an indicator function defined by
			\vspace{-7pt}
			\begin{equation} \label{eq:indicator_function}
			    x(i, j) = 
			    \begin{cases}
			    1, & \text{if } i < j \text{ and } \{i, j\} \in (\mathcal{E} \cup \mathcal{Q}) \\
			    0 & \text{otherwise}
			    \end{cases}
			\vspace{-7pt}
			\end{equation}
			and $\mathbf{e}_{\Delta \mathbf{T}}$ denotes the relative pose residuals and $\mathbf{W}_{\Delta \mathbf{T}}$ the information matrix of the relative pose constraints (\refsec{sec:meth:residuals}).
			The sets $\mathcal{E}$ and $\mathcal{Q}$ denote the covisibility edges between \acp{KF} and loop closure edges, respectively.
			After the optimization, the positions of all \acp{LM} in the server map are propagated using the optimized \ac{KF}-poses. 
		\subsubsection{Global Bundle Adjustment}
			COVINS performs on-demand \ac{GBA}, e.g. at the end of the mission when the agents are not actively sending further information to the server. 
			This creates a highly accurate estimate, which can be re-used in a multi-session fashion for further collaborative SLAM session.
			For a specific server map, we perform \ac{GBA} taking into account all \acp{KF} and \acp{LM} in the map, using the following objective function:
			\vspace{-5pt}
			\begin{align} \label{eq:global_bundle_adjustment}
            	J_{GBA}(\bm{\Theta}) &\coloneqq \lVert \bm{e}_{p}^{c}\rVert^{2}_{\bm{W}_{p}^{c}}  + \sum\limits_{k \in \mathcal{V}} \sum\limits_{j \in \mathcal{L}(k)}
            	\delta\left(\lVert \bm{e}_{r}^{k,j} \rVert^{2}_{\bm{W}_{r}^{k,j}} \right)  \\
            	&+ \sum\limits_{k-1, k \in \mathcal{V}} \left( \lVert \bm{e}_{IMU}^{k-1,k} \rVert^{2}_{\bm{W}_{IMU}^{k-1,k}} +  \lVert \bm{e}_{b}^{k-1,k} \rVert^{2}_{\bm{W}_{b}^{k-1,k}}  \right)  \nonumber ~,
        	\end{align}
        	where the first term corresponds to a prior added to the first \ac{KF} in order to remove the Gauge degree of freedom,  and $\mathcal{L}(k)$ denotes the set of \acp{LM} observed by $\textit{KF}_{k}$.
        	The function $\delta(\cdot)$ denotes the use of a robust cost function to reduce the influence of outlier observations, in our case the Cauchy loss is used.
        	The terms $\mathbf{e}_{r}$ and $\mathbf{e}_{\text{IMU}}$ correspond to the reprojection and IMU pre-integration residuals (\refsec{sec:meth:residuals})
        	The term $\mathbf{e}_{b}^{k-1,k}$ penalizes changes in the bias variables between successive \acp{KF}.
        	After the optimization, a outliers with large reprojection residuals are removed from the map.
			Note that the IMU constraints in \refeq{eq:global_bundle_adjustment} are only inserted between consecutive KFs created by the same agent.
		\subsubsection{Redundancy Removal (KF Pruning)} \label{sec:meth:opt:red}
			Creating a large number of \acp{KF} is beneficial for \ac{VIO} to achieve a high level of robustness and accuracy.
			However, for the global \ac{SLAM} estimate, an increasing number of \acp{KF} results in increasing runtime of the employed algorithms, notably of the optimization algorithms, scaling cubic with the number of \acp{KF} in the worst case.
			Therefore, it is desirable to remove redundant \acp{KF} from the \ac{SLAM} graph to increase scalability of the system.
			For this reason, we employ the structure-based heuristic introduced in \cite{schmuck2019:3DV:redundancy} to identify and remove redundant \acp{KF}.
			The underlying assumption of the heuristic is that with increasing number of observations of a specific \ac{LM} (by different \acp{KF}), the information gained by an individual observation decreases.
			Therefore, with $\text{obs}(\textit{LM}_i)$ denoting the number of observations of $\textit{LM}_i$, a function $\tau(x):\mathbb{N}^0 \to [0,1]$ assigns a value to each \ac{LM} depending on its number of observations, with increasing number encoding increasing redundancy of an individual observation of this \ac{LM}.
			The complete definition of $\tau$ can be found in \cite{schmuck2019:3DV:redundancy,covins:supp}.
			Using $\mathcal{L}(k)$, the set of \acp{LM} observed by $\textit{KF}_{k}$, we can  calculate a redundancy value $\phi(\cdot)$ for each $\textit{KF}_k$ in the map as
			\vspace{-7pt}
			\begin{equation}
			\phi(\textit{KF}_{k}) = \frac{1}{|\mathcal{L}(k)|} \sum_{i \in \mathcal{L}(k)} \tau(\text{obs}(\textit{LM}_i))
			\label{eq:mth:struct:phi}
			\vspace{-5pt}
			\end{equation}
			This way, assigning a value to each \ac{KF} estimating its information contributed to the SLAM estimate, the most redundant \acp{KF} can be removed from the estimate.
            Redundancy removal is performed before \ac{GBA}, since the timing of \ac{GBA} is affected the strongest by the number of \acp{KF}.
			\textbf{\ac{LM} pruning} is implicitly handled: whenever a \ac{LM} becomes under-observed (i.e. less than 2 observations), either from removal of outlier observations or through removing \acp{KF}, this \ac{LM} is removed from the map.
\section{Experimental Results} \label{sec:exp}
	We evaluate COVINS in a thorough testbed of experiments, investigating its accuracy using the EuRoC benchmark dataset \cite{dataset:burri2016euroc} employing a local PC as well as an \ac{AWS} cloud server (\refsec{sec:exp:cslam_euroc}), scalabilty in large-scale experiments with 12 agents (\refsec{sec:exp:cslam_12}), drift correction (\refsec{sec:exp:drift}), the influence of the redundancy removal (\refsec{sec:exp:red}) and communication statistics (\refsec{sec:exp:comm}).
	All results were obtained by re-playing data in real-time, and values in this section are averaged over 5 runs for each experiment if not stated otherwise.
	For these experiments, the following setup is used:
	\vspace{-5pt}
	\begin{itemize}
	    \setlength\itemsep{-0.3em}
		\item Local Server: Lenovo T480s (1.80 GHz $\times$ 8 (max 4.00 GHz))
		\item Cloud Server: AWS c5a.8xlarge (32 vCPUs at 3.3 GHz)
        \item Agents: Intel NUC 7i7BNH with 3.5 GHz $\times$ 4
    \vspace{-5pt}
	\end{itemize}
	Throughout all experiments, the pre-recorded datasets are processed onboard the agents, which are connected to the server via a wireless network, so that real communication takes place.
	This makes our evaluation across different runs more comparable and provides us with ground truth, while still using real network communication as it would be the case during a real-world application.
	\subsection{Collaborative SLAM Estimation Accuracy} \label{sec:exp:cslam_euroc}
\begin{figure}[t]
			\centering
            \includegraphics[width=0.99\columnwidth]{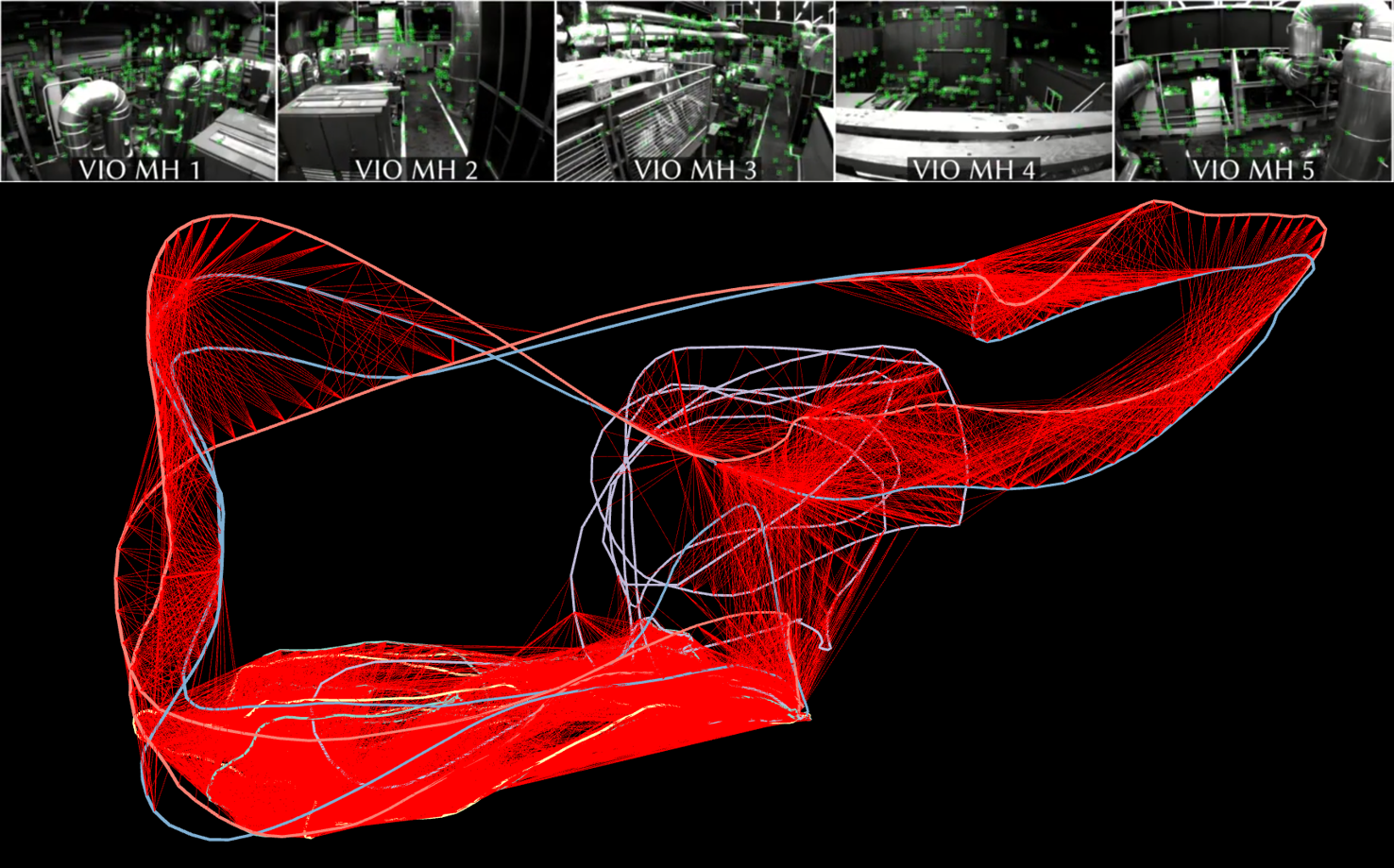}
            \vspace{-5pt}
			\caption{
				Collaborative \ac{SLAM} estimate with five agents using EuRoC MH1-MH5. 
                Top: Agent scene views.
			}
			\label{fig:ex:coll_euroc}
			\vspace{-15pt}
		\end{figure}

	    We evaluate the accuracy of the global collaborative SLAM estimate of COVINS using \cite{dataset:burri2016euroc}, where we use the five Machine Hall (MH) sequences, and the three Vicon Room 1 (V1) sequences to establish a collaborative estimation scenario with three and five participating agents.
	    A global estimate jointly created by five agents is shown in \reffig{fig:ex:coll_euroc}.
	    \reftab{tab:ex:coll_euroc} reports the accuracy of the aligned global estimate in terms of \ac{ATE} and scale error, as well as a comparison to ORB-SLAM3 \cite{campos2021:TRO:orb3}, VINS-mono \cite{qin2018:TRO:vins} (both having multi-session functionalities) and CVI-SLAM \cite{karrer2018cvi} using the \textit{Local Server} for all experiments.
		\begin{table}[b]
			\vspace{-5pt}
			\renewcommand{\arraystretch}{1.0}
			\caption{
				Multi-agent map evaluation (\ac{ATE} in $m$, scale error in \%), using \cite{dataset:burri2016euroc} (lowest error in bold).
				The front-end of \cite{karrer2018cvi} is not able to track the highly dynamic motions of the V10\textit{x} sequences, therefore, no values are reported.
                Last row: results on the AWS cloud server.
			}
			\vspace{-5pt}
			\label{tab:ex:coll_euroc}
			\centering
			\scalebox{0.88}{
				\begin{tabular}{|c|c|c|c|}
					\hline
					\bfseries \multirow{2}{1.1cm}{System} & \multicolumn{3}{c|}{\bfseries Sequences} \\
					\cline{2-4}
					\bfseries & MH01-MH03 & MH01-MH05 & V101-V103 \\			
					\hline\hline
					CVI-SLAM \cite{karrer2018cvi} & \makecell{0.054 (0.47\%)} &  \makecell{0.091 (1.02\%)} &  \makecell{--- (---)} \\
					\hline
					ORB-SLAM3 \cite{campos2021:TRO:orb3} & \makecell{0.041 (2.21\%)} &  \makecell{0.082 (1.13\%)} &  \makecell{0.048 (1.30\%)} \\
					\hline
					VINS-mono \cite{qin2018:TRO:vins} &  \makecell{{0.062} (0.31\%)} &  \makecell{{0.100} (\textbf{0.08\%})} & \makecell{{0.076} (1.34\%)} \\
					\hline
					COVINS &  \makecell{\textbf{0.024} (\textbf{0.24\%})} &  \makecell{\textbf{0.036} (0.3\%)} &  \makecell{\textbf{0.042} (\textbf{1.01\%)}} \\
					\hline
					\hline
					COVINS AWS &  \makecell{0.025 (0.36\%)} &  \makecell{0.039 (0.42\%)} &  \makecell{0.049 (1.81\%)} \\
					\hline
				\end{tabular}
			}
		\end{table}
        COVINS shows generally high accuracy across all datasets, achieving similar or better performance than the  state of the art.
	    The high quality of COVINS' estimate in multi-agent scenarios is due to the fact that the framework is able to establish a large number of accurate constraints between the data contributed by the individual agents, as visible from \reffig{fig:ex:coll_euroc}, where the red lines encode covisibility edges between separate trajectories. 
	    Furthermore, \reftab{tab:ex:coll_euroc} reports the results for the same experimental setup for COVINS, except that an AWS cloud server is now used to run the server back-end of COVINS.
	    The accuracy of this cloud-based estimation collaborative SLAM estimate is similar to the accuracy using a local server, attesting to the capability of the COVINS back-end to be executed on remote cloud compute, with potentially much higher computational resources than a locally deployed PC.
	\subsection{Large-Scale Collaborative SLAM with 12 Agents} \label{sec:exp:cslam_12}
		In this experiment, we evaluate the applicability of COVINS to a large-scale scene and a large team of participating agents.
        For this, we use a newly generated dataset with 12 UAVs equipped with a downward looking camera flying over a small village.
        In order to obtain accurate ground truth, the dataset was created using the visual-inertial simulator from \cite{teixeira2020:RAL:aerial}, creating photo-realistic vision datasets for UAV flights using a high-quality 3D model of the scene.
        It comprises 12 circular UAV trajectories of $20m$ radius, covering an area of about $500m^2$ with $1750m$ total trajectory length.
		\reffig{fig:intro:intropic} shows the final collaborative estimate generated by COVINS, consisting of
		over 3200 \acp{KF} and about 200k \acp{LM}.
		The average \ac{ATE} of the estimate is $0.050 m$, the average scale error is 0.44\%.
		An illustration of the 3D scene  is shown in \cite{covins:supp} and the accompanying video
		\footnote{\url{https://youtu.be/FxJTY5x1fGE}}.
		%
	\subsection{Drift Correction} \label{sec:exp:drift}
\begin{figure}[t]
			\centering
			\begin{subfigure}[thpb]{0.33\textwidth}
				\includegraphics[height=3.3cm]{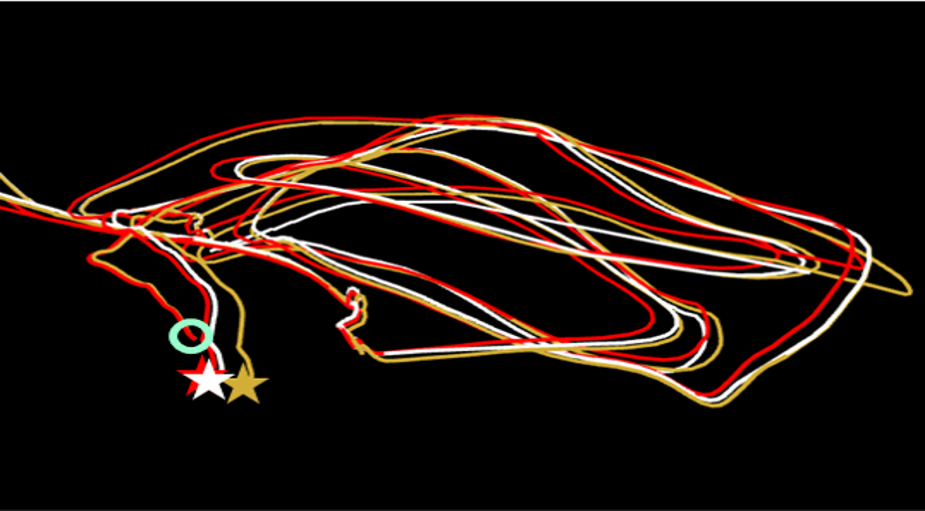}
				\vspace{-15pt}
				\caption{Final \ac{VIO} estimated location (gold) and drift-corrected estimate (white) onboard the agent on EuRoC MH3. 
				}
				\label{fig:ex:drift_corr_mh3}
			\end{subfigure}
			\hspace{2pt}
			\begin{subfigure}[thpb]{0.13\textwidth}
				\includegraphics[height=3.3cm]{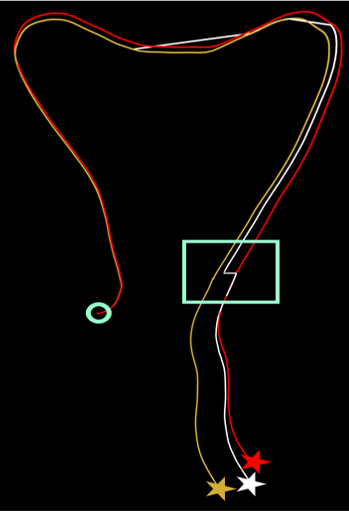}
				\vspace{-15pt}
				\caption{
				Drift-correction after loop closure (MH5).
				}
				\label{fig:ex:drift_corr_mh5}
			\end{subfigure}
			\vspace{-5pt}
			\caption{
			Drift correction, with ground truth (red), \ac{VIO} trajectory estimate (gold), and drift-corrected estimate (white).
			The star indicates the agent's current position. The green circle marks the origin.
			}
			\label{fig:ex:drift_corr}
			\vspace{-15pt}
		\end{figure}

		\reffig{fig:ex:drift_corr} visually demonstrates the effect of the drift estimation and correction scheme.
		\reffig{fig:ex:drift_corr_mh3} shows the final trajectory estimated by the agent's onboard \ac{VIO} system using the MH3 sequence.
        Note that although the full trajectory is displayed, this trajectory was never globally optimized by the agent itself.
		As visible from \reffig{fig:ex:drift_corr_mh3}, the estimated trajectory (gold) is affected by some drift, so that the estimated final location of the agent deviates from the true location (red).
		However, based on the continuous estimation of the drift using the information received from the server, the agent can estimate a corrected trajectory (white), being noticeably closer to the true trajectory.
		A similar effect can be seen from \reffig{fig:ex:drift_corr_mh5}, which reflects a snapshot of the \ac{VIO} estimate during an experiment using the MH5 sequence, with the green box highlighting a significant drift correction with information received from the server after loop closure.
		\subsection{Redundancy Detection \& Removal} \label{sec:exp:red}
		Using the maps created during the 5-agent experiments in \refsec{sec:exp:cslam_euroc}, we evaluate the influence of the redundancy detection scheme implemented in COVINS.
		The evaluation is performed as follows: the five multi-agent maps created during the five 5-agent experiments on MH1-MH5 of \refsec{sec:exp:cslam_euroc} were saved to file storage after each experiment, so that they can be reloaded into COVINS for further experiments.
		For each map, which contains on average approximately 1700 \acp{KF} each, a reduction of the map to 1250, 1000 and 750 \acp{KF} is performed in separate experiments.
		The results (average \ac{ATE} over all five maps) are reported in \reftab{tab:ex:red_euroc}, demonstrating that COVINS is able to significantly reduce the number of \acp{KF} in the estimate at only a small loss in accuracy and a significant reduction of the \ac{GBA} time.
		Even when compressing the map by more than 50\%, the mean error increases by only $7 mm$.
		\begin{table}[thpb]
			\vspace{-5pt}
			\renewcommand{\arraystretch}{1.3}
			\caption{
				Evaluation of redundancy detection on the map accuracy.
			}
			\vspace{-5pt}
			\label{tab:ex:red_euroc}
			\centering
			\scalebox{0.90}{
				\begin{tabular}{|c|c|c|c|c|}
					\hline
					\bfseries Num. KFs & 1681 (init. state) & 1250 & 1000 & 750  \\
					\hline
					\bfseries ATE [m] & 0.036 & 0.039 & 0.04 & 0.043 \\
					\hline
					\bfseries GBA Time [s] & 133 & 53 & 34 & 20 \\
					\hline
				\end{tabular}
			}
			\vspace{-10pt}
		\end{table}
	\subsection{Communication Statistics} \label{sec:exp:comm}
		\reftab{tab:ex:comm_traffic} reports the network traffic generated by the communication between sever and agent. 
		Each agent informs the server at a frequency of 5 Hz about new and modified map data since the previously sent data bundle.
        The server shares information with the agent at a 2 Hz frequency.
        As visible from \reftab{tab:ex:comm_traffic}, the generated network traffic from an individual agent to the server lies approximately between 400 and 600 KB/s, which can comfortably be covered by typical WiFi infrastructure.
        More challenging sequences require \ac{VIO} to create more \acp{KF} for successful operation, translating to more transmitted data (e.g. MH1 (easiest): 2.9 \acp{KF}/s created, V103 (hardest): 6.9 KFs/s). 
        The traffic from the server to the individual agent is significantly lower in our implementation, since only pose information for a single \ac{KF} needs to be shared for the drift correction scheme.
        The average size of the individual message types is as follows: \ac{KF} full: 97kB; \ac{KF} update: 273 byte; \ac{LM} full: 162 byte; \ac{LM} update: 65 byte.
        \begin{table}[thpb]
        	\vspace{-5pt}
        	\renewcommand{\arraystretch}{1.3}
        	\caption{
        		Network traffic of agent-to-server and vice-versa, and total time consumed for communication onboard the agent, averaged over all 8 EuRoC sequences used, and for selected individual sequences.
        	}
        	\vspace{-5pt}
        	\label{tab:ex:comm_traffic}
        	\centering
        	\scalebox{0.90}{
        		\begin{tabular}{|c|c|c|c|}
        			\hline
        			\bfseries Sequence & \bfseries Agent $\to$ Server & \bfseries Server $\to$ Agent & \bfseries Comm Time \\
        			\hline\hline
        			Avg. (8 seq.) & 493.36 kB/s & 2.31 kB/s & 792.91 ms \\
        			\hline\hline
        			MH1 & 422.83 kB/s & 2.29 kB/s & 939.92 ms \\
        			\hline
        			MH5 & 540.37 kB/s & 2.32 kB/s & 776.33 ms \\
        			\hline
        			V103 & 609.22 kB/s & 2.31 kB/s & 850.24 ms \\
        			\hline
        		\end{tabular}
        	}
        	\vspace{-8pt}
        \end{table}
        \reftab{tab:ex:comm_traffic} also reports the timings of the communication module on the agent.
        With approximately 1s of total communication time for each sequence, and the sequences containing flights between 84s (V102) and 144s (V101), the overhead of the communication is marginal ($< 1\%$) compared to the total time of the estimation process and does not compromise the real-time capability of the \ac{VIO} system.
\section{Conclusion}
	In this paper, we present COVINS, a powerful and accurate back-end for collaborative SLAM.
	COVINS allows multiple agents to generate collaborative global SLAM estimates from their simultaneously contributed data online during the mission, eliminating the need for external infrastructure or pre-built maps in order to enable multi-agent applications.
	The efficient architecture and system design of COVINS allows this framework to process data contributed by many agents simultaneously.
	Our experiments attest to the high accuracy of the collaborative SLAM estimates in large-scale multi-agent missions, in particular demonstrating collaborative SLAM with up to 12 agents contributing to the system, which, to the best of our knowledge, is the highest number of participants demonstrated by any comparable system in the literature.
	Boosting applicability and scalability of the system, this framework can run locally on a PC as well as on a remote cloud server, furthermore, supported by a redundancy detection scheme that was demonstrated to be a able to significantly reduce the number of \acp{KF} in the estimate, while keeping a similar level of accuracy.
	Future work will focus on further leveraging the applicability and scalability of the system to potentially hundreds of agents, and interfacing COVINS with a front-end that is able to run on mobile devices, such as VINS-Mobile \cite{qin2018:TRO:vins}, furthermore enabled to display \acs{AR} content, in order to leverage COVINS' collaborative scene understanding to create and demonstrate a shared \acs{AR} experience for multiple users.
\bibliographystyle{abbrv-doi}
\bibliography{main}
\end{document}



\firstsection{Notation} \label{sec:prelim:notation}
\maketitle
    In this work, we adopt the notation from \cite{karrer2018cvi} for mathematical notation and 
    state and error residual formulations. 
    Small bold letters (e.g. $\mathbf{a}$) and large bold letters (e.g. $\mathbf{A}$) denote vectors and matrices, respectively. 
    Coordinate frames are denoted as plain capital letters (e.g. $A$).
    In order to denote a vector $\mathbf{x}$ expressed in $A$, the notation $\prescript{}{A}{\mathbf{x}}$ is used.
    A rigid body transformation transforming a point from coordinate frame $B$ to $A$ is denoted as $\mathbf{T}_{AB}$, where the translational and rotational part of any $\mathbf{T}$ is denoted by $\mathbf{t}$ and $\mathbf{R}$, respectively.
    Throughout this work we denote the world frame (i.e. the inertial frame) as $W$, the \ac{IMU} body frame as $S$ and the camera frame as $C$.
\section{IMU Model and System States} \label{sec:prelim:state}
    In order to incorporate \ac{IMU} information into COVINS, we model the \ac{IMU} using a standard model.
    Assuming that measurements from both the acceleration $_{S}\mathbf{a}(t)$ and the gyroscope $_{S}\pmb{\omega}_{WS}(t)$ are corrupted by additive Gaussian noise $\mathbf{\eta}$ and have an unknown, time varying sensor bias, the \ac{IMU} readings are modelled as follows:
    \begin{align} \label{eq:imu_model}
    \prescript{}{S}{\mathbf{a}}(t) &= \mathbf{R}_{WS}^{T}(t)(\prescript{}{W}{\hat{\mathbf{a}}}(t) - \prescript{}{W}{\mathbf{g}}) + \mathbf{b}_{a}(t) + \pmb{\eta}_{a}(t),  \\
    \prescript{}{S}{\pmb{\omega}}_{WS}(t) &= \prescript{}{S}{\hat{\pmb{\omega}}}_{WS}(t) + \mathbf{b}_{g}(t) + \pmb{\eta}_{g}(t) ~.
    \end{align}
    The true values for the variables are indicated by a $\hat{\cdot}$ and $_{W}\mathbf{g}$ denotes the gravity vector.
    To account for this \ac{IMU} model, the system state $\pmb{\Theta}$ includes besides the \ac{KF} poses and \ac{LM} positions also the linear velocities $_{W}\mathbf{v}$ as well as the bias variables for each \ac{KF} $k$:
    \begin{equation}\label{eq:system_state}
    \pmb{\Theta} \coloneqq \{ \underbrace{\mathbf{R}_{WS}^{k}, \mathbf{t}_{WS}^{k}, _{W}\mathbf{v}^{k}, \mathbf{b}^{k}}_{\text{KF}_{k}}, \prescript{}{W}{\mathbf{l}}^{i} \}, \forall k \in \mathcal{V}, \forall i \in \mathcal{L} ~,
    \end{equation}
    where the sets $\mathcal{V}$ and $\mathcal{L}$ denote the set of all \acp{KF} and \acp{LM}, respectively.
    Whenever the context allows for it, we use $\theta_{j}$ to denote an individual state variable.
\section{Error Residuals Formulation } \label{sec:meth:residuals}
    By formulating a set of residuals, the optimization of state variables occurring in \ac{KF}-based VI SLAM can be expressed as a weighted nonlinear least-squares problem.
    Each such residual $\mathbf{e}_{i}$ expresses the difference between the expected measurement based on the current state of the system and the actual measurement $\mathbf{z}_{i}$:
    \begin{equation}\label{eq:residual_form}
        \mathbf{e}_{i} \coloneqq \mathbf{z}_{i} - h_{i}(\mathcal{A}_{i}) ~,
    \end{equation}
    where $\mathcal{A}_{i}$ is the set of state variables $\theta_{j}$ relevant for measurement $\mathbf{z}_{i}$, and $h_{i}(\cdot)$ is the measurement function, predicting the measurement according to these state variables in $\mathcal{A}_{i}$.
    %
    By collecting all occurring residual terms, the objective of the optimization can be expressed as:
    \begin{equation}
        \pmb{\Theta}^{*} = \underset{\pmb{\Theta}}{\mathrm{arg min}} \left\{ \sum\limits_{i} \lVert \mathbf{z}_{i} - h_{i}(\mathcal{A}_{i}) \rVert_{\mathbf{W}_{i}}^{2} \right\} ~,
    \end{equation}
    where $\lVert \mathbf{x} \rVert_{W}^{2} = \mathbf{x}^{T} \mathbf{W} \mathbf{x}$ denotes the squared Mahalanobis distance with the information matrix $\mathbf{W}$.
    Within our system, we essentially use three different types of residuals:
    reprojection residuals $\mathbf{e}_{r}$, relative pose residuals $\mathbf{e}_{\Delta \mathbf{T}}$, and IMU pre-integration residuals $\mathbf{e}_{\text{IMU}}$.
    \subsection{Reprojection Residual}
    The reprojection residual relates the projection of \ac{LM} j, with its position $_{W}\mathbf{l}^{j}$, to the associated 2D image observation $\mathbf{z}^{k,j}$ in \ac{KF} $k$ and is given by:
    \begin{equation} \label{eq:repojection_error}
        \mathbf{e}_{r}^{k,j} \coloneqq \mathbf{z}^{k,j} - \pi \left(\mathbf{T}_{CS}^{k} (\mathbf{T}_{WS}^{k})^{-1} _{W}\mathbf{l}^{j}\right) ~.
    \end{equation}
    \subsection{Relative Pose Residual}  \label{sec:app:residuals:relpose}
    The relative pose residuals are used to form constraints across two poses $\mathbf{T}_{i} (=\mathbf{T}_{RI})$  and $\mathbf{T}_{j} (=\mathbf{T}_{RJ})$, with $R$ being a reference frame.
    The measurement $\Delta\mathbf{T}_{ij}$ provides an estimate of the transformation between the frames $I$ and $J$.
    Using that, the residual can be formulated as
    \begin{equation}\label{eq:relative_pose_residual}
        \mathbf{e}_{\Delta \mathbf{T}} \coloneqq \begin{bmatrix}
                \log\left( \Delta\mathbf{R}_{ij} \mathbf{R}_{j}^{T} \mathbf{R}_{i} \right)^{T} &  
                \left( \Delta\mathbf{t}_{ij} + \mathbf{R}_{j}^{T} \mathbf{t}_{j} - \mathbf{t}_{i} \right)^{T}
        \end{bmatrix}^{T} ~,
    \end{equation}
    where the $\log\left(\mathbf{R}\right)$ defines the mapping of a 3D rotation to its tangent plane, as described in \cite{bloesch2016_rotation}.
    \subsection{IMU pre-integration Residual}
    Given a sequence of \ac{IMU} readings, both accelerometer and gyroscope, between two \acp{KF}, a relative constraint between the two \acp{KF} can be obtained by the means of integration of the \ac{IMU} measurements.
    However, as described in \refeq{eq:imu_model}, these measurements are affected by an unknown bias, which in return influences the result of the integration.
    In order to avoid the costly re-integrating all \ac{IMU} measurements upon every change of the bias variables (i.e. after every optimization step), \cite{forster2017manifold} presented a method allowing to perform the integration only once and apply linearized corrections considering the bias changes w.r.t. the bias estimate used at the time of the integration.
    Utilizing this pre-integration scheme, the resulting \ac{IMU} residual terms can be written as:
    \resizebox{1.0\linewidth}{!}{
	\begin{minipage}{\linewidth}
    \begin{align}
		\label{eq:preintegration_parts}
		\bm{e}_{\Delta \bm{R}}^{k-1,k} &= \log \left( \left(\Delta \tilde{\bm{R}}_{k-1,k}(\bar{\bm{b}}_{g}^{k-1}) \exp\left(\frac{\partial \Delta \bar{\bm{R}}_{k-1,k}}{\partial \bm{b}_{g}} \delta \bm{b}_{g} \right)  \right)^{\intercal} \right. \nonumber \\ 
		& \left. \bm{R}_{WS}^{k-1^{\intercal}} \bm{R}_{WS}^{k} \right) \nonumber \\
		\bm{e}_{\Delta \bm{v}}^{k-1,k} &= \Delta \bm{R}_{WS}^{k-1^{\intercal}} \left( {}_{W}\bm{v}^{k} - {}_{W}\bm{v}^{k-1} - {}_{W}\bm{g}\Delta t_{k-1,k} \right)   \nonumber\\
		&- \left(\Delta \tilde{\bm{v}}_{k-1,k}(\bar{\bm{b}}) + \frac{\partial \Delta \bar{\bm{v}}_{k-1, k}}{\partial \bm{b}_{a}} \delta\bm{b}_{a} + \frac{\partial \Delta \bar{\bm{v}}_{k-1,k}}{\partial \bm{b}_{g}} \right) \\
		\bm{e}_{\Delta\bm{t}}^{k-1,k} &= \bm{R}_{WS}^{k-1^{\intercal}} \left(
		\Delta \bm{t}_{WS}^{k} - \bm{t}_{WS}^{k-1} - {}_{W}\bm{v}^{k-1} \Delta t_{k-1,k} - \frac{1}{2}\bm{g} \Delta t_{k-1,k}^{2}\right) \nonumber \\
		&-\left(\Delta \tilde{\bm{t}}_{k-1,k} (\bar{\bm{b}}^{k-1}) + \frac{\partial \Delta \bar{\bm{t}}_{k-1,k}}{\partial  \bm{b}_{a}}\delta \bm{b}_{a} + \frac{\partial \Delta \bar{\bm{t}}_{k-1,k}}{\partial \bm{b}_{g}} \delta \bm{b}_{g} \right) ~, \nonumber
	\end{align}
	\end{minipage}
    }
    where the values indicated $\bar{\cdot}$ denote variables obtained with the bias estimate $\bar{\mathbf{b}}$ at the time of the pre-integration and values with $\tilde{\cdot}$ indicate values using the current estimate of the state variables.
    The scalar value $\Delta t_{k-1,k}$ represents the time interval between the two \acp{KF} at time $k-1$ and $k$.
    The individual error terms in \refeq{eq:preintegration_parts} are put together into a residual vector as
    \begin{equation} \label{eq:preintegration_residual}
        \mathbf{e}_{\text{IMU}}^{k-1,k} \coloneqq 
        \begin{bmatrix}
                \left. \bm{e}_{\Delta \bm{R}}^{k-1,k} \right.^T & \left.\bm{e}_{\Delta \bm{v}}^{k-1,k}\right.^{T} & \left. \bm{e}_{\Delta\bm{t}}^{k-1,k} \right.^{T}
        \end{bmatrix}^{T}
    \end{equation}
\section{Factor Graph Representation of the Map Structure} \label{sec:meth:map}
\begin{figure}[t]
		\centering
		\includegraphics[scale=0.5]{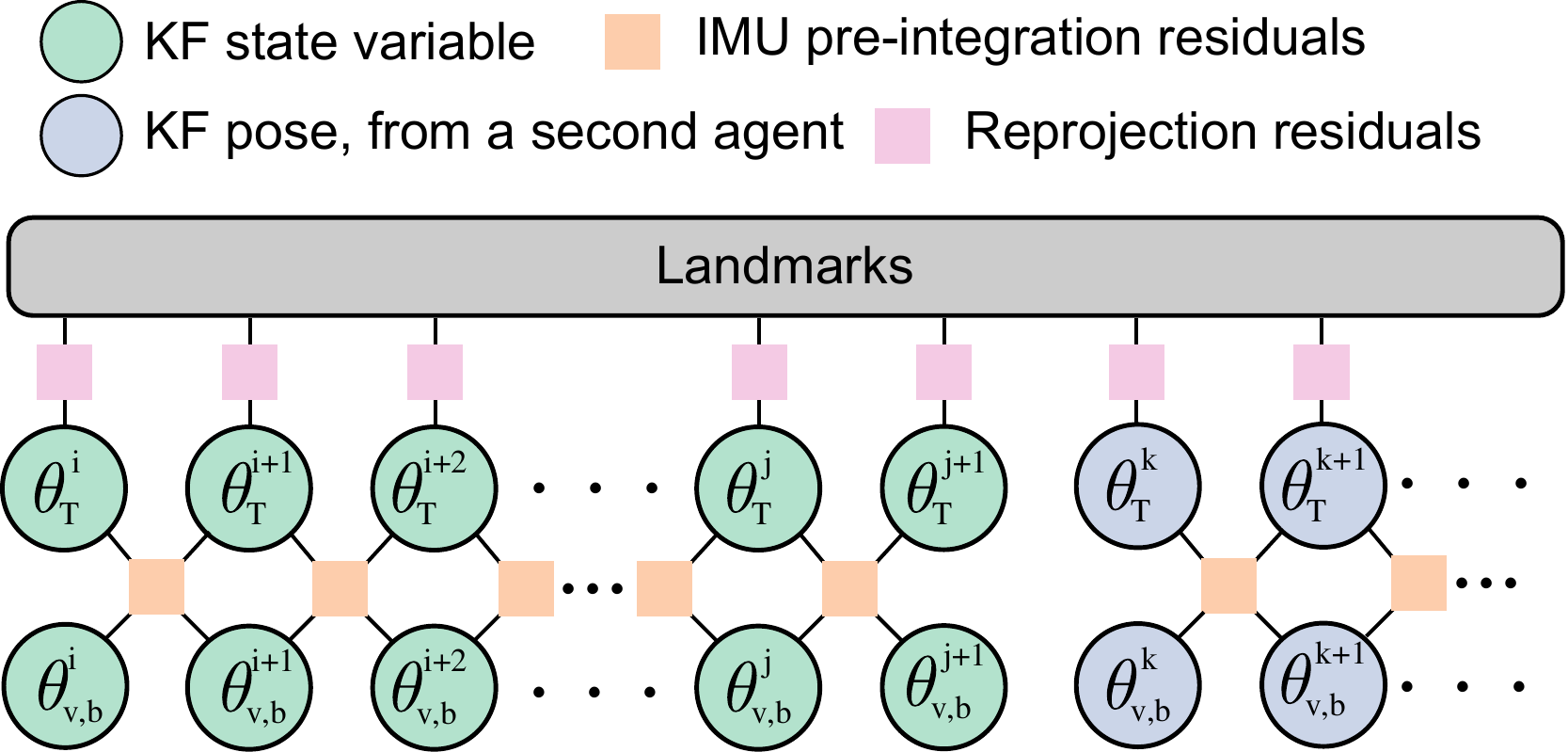}
		\caption{
			Schematic depiction of the factor graph formulation of the \ac{SLAM} map with all relevant variables, which  are involved in \ac{GBA}. \acp{KF} captured by different agents get associated via mutual observations of \acp{LM}, while IMU residuals are only connecting \acp{KF} from the same agent. 
			Note: possible additional constraints from loop closures are not illustrated here.
		}
		\label{fix:meth:factorgraph}
	\end{figure}

	Together with the state definition (\refsec{sec:prelim:state}) and the error residuals (\refsec{sec:meth:residuals}), the underlying \acs{SLAM} estimation problem of COVINS induces a factor graph, illustrated in \reffig{fix:meth:factorgraph}, which forms the basis of the \ac{GBA} scheme employed in the framework.
	As shown, shared \ac{LM} observations create dependencies between \acp{KF} from multiple agents, while \ac{IMU} factors are only inserted between consecutive \acp{KF} created by the same agent.
\section{Observation Redundancy Measure} \label{sec:app:obsred}
    Following \cite{schmuck2019:3DV:redundancy}, the function $\tau(x):\mathbb{N}^0 \to [0,1]$ assigns a value to each \ac{LM} depending on its number of observations $x$, with increasing number encoding increasing redundancy of an individual observation of this \ac{LM}.
    In conformity with \cite{schmuck2019:3DV:redundancy}, with a minimum ob two observations required to constrain a \ac{LM}, $\tau(2) = 0$, and $\tau(x) = 1$ for $x > 5$, so that five observations are considered sufficient to robustly estimate the 3D position of a \ac{LM}.
    Taking into account the aspect that the redundancy of one individual observation increases  with increasing number of observations, $\tau(x)$ is defined as follows:
    \begin{equation}
    \tau(x) =     
    \begin{cases}
    0, & \text{if}\ x \leq 2 \\
    0.4, & \text{if}\ x = 3 \\
    0.7, & \text{if}\ x = 4 \\
    0.9, & \text{if}\ x = 5 \\
    1, & \text{if}\ x > 5 \\
    \end{cases}
    \end{equation}
\section{Experimental Results} \label{sec:exp}
	\subsection{Large-Scale Collaborative SLAM with 12 Agents} \label{sec:exp:cslam_12}
\begin{figure}[t]
			\centering
			\begin{subfigure}[thpb]{0.19\textwidth}
				\includegraphics[height=3.5cm]{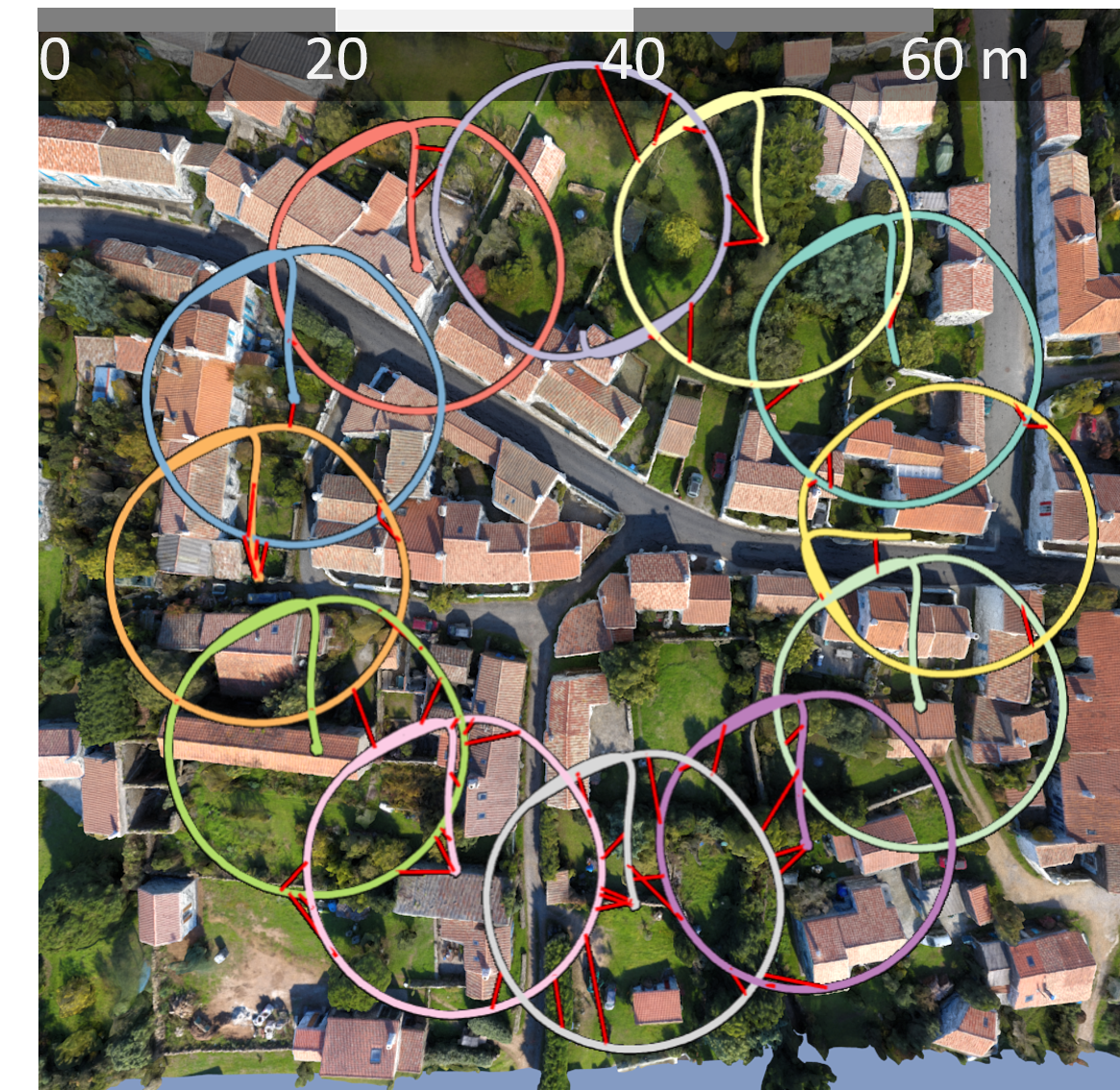}
				\caption{Final 12-agent collaborative \ac{SLAM} estimate (trajectories only), superimposed on the scene view. \newline
				}
				\label{fig:ex:village_scene}
			\end{subfigure}
			\hspace{3pt}
			\begin{subfigure}[thpb]{0.26\textwidth}
				\includegraphics[height=3.5cm]{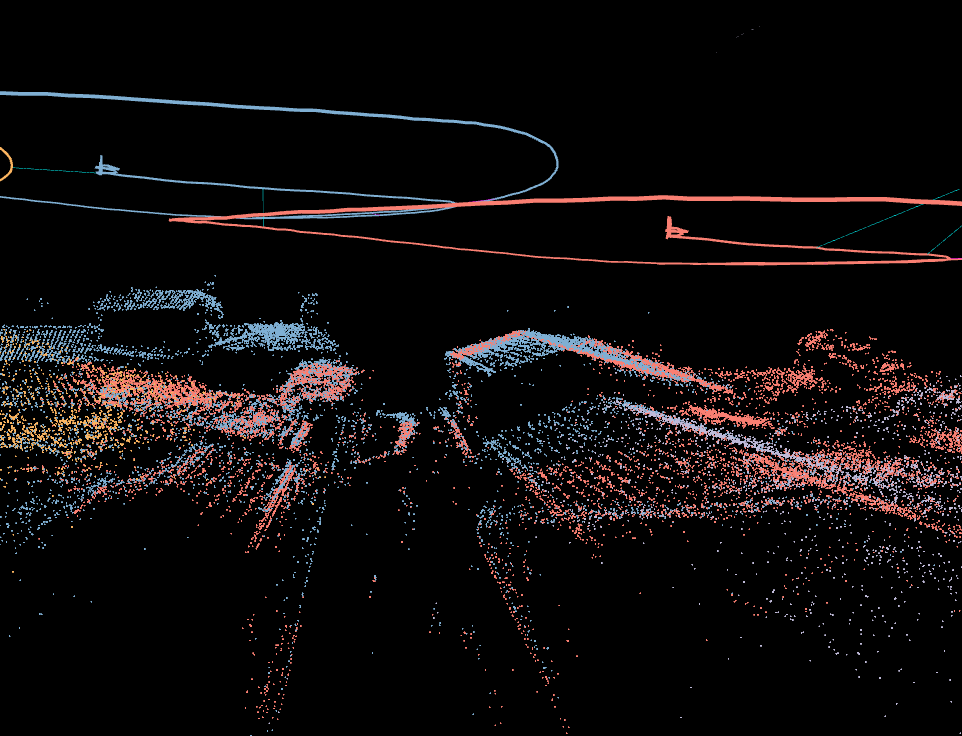}
				\caption{Side view on the trajectories and \ac{LM} at the intersection between two trajectories, illustrating the accurate alignment of individual map data.}
				\label{fig:ex:village_zoom}
			\end{subfigure}
			\caption{The collaborative \ac{SLAM} estimates in the 12-agent village experiment.}
			\label{fig:ex:village}
		\end{figure}

        \reffig{fig:ex:village_scene} illustrates a scene view for the \textit{village dataset} used for the 12-agent experiment reported in COVINS, as well as the 12 circular UAV trajectories of $20m$ radius.
		\reffig{fig:ex:village_zoom} shows a side view on the sparse \ac{LM} point cloud on the intersection between two trajectories, attesting to the high accuracy of the collaborative estimate through the well-aligned \acp{LM} of the two agents.
	\subsection{Communication Statistics} \label{sec:exp:comm}
		\reftab{tab:ex:comm_traffic} reports the network traffic generated by the communication between sever and agent for all sequences of the EuRoC dataset \cite{dataset:burri2016euroc}, including those that are omitted in the main paper.
        \begin{table}[thpb]
        	\renewcommand{\arraystretch}{1.3}
        	\caption{
        		Network traffic of agent-to-server and vice-versa, and total time consumed for communication onboard the agent (5-run-average). The first row shows the average over all 8 different EuRoC sequences.	The generation of more \acp{KF} onboard the agent for the more challenging sequences leads to increasing network traffic.
        	}
        	\label{tab:ex:comm_traffic}
        	\centering
        	\scalebox{0.95}{
        		\begin{tabular}{|c|c|c|c|}
        			\hline
        			\bfseries Sequence & \bfseries Agent $\to$ Server & \bfseries Server $\to$ Agent & \bfseries Comm Time \\
        			\hline\hline
        			Avg. & 493.36 kB/s & 2.31 kB/s & 792.91 ms \\
        			\hline\hline
        			MH1 & 422.83 kB/s & 2.29 kB/s & 939.92 ms \\
        			\hline
        			MH2 & 450.78 kB/s & 2.31 kB/s & 825.37 ms \\
        			\hline
        			MH3 & 463.82 kB/s & 2.31 kB/s & 771.50 ms \\
        			\hline
        			MH4 & 554.26 kB/s & 2.33 kB/s & 652.41 ms \\
        			\hline
        			MH5 & 540.37 kB/s & 2.32 kB/s & 776.33 ms \\
        			\hline
        			V101 & 402.29 kB/s & 2.29 kB/s & 1063.57 ms \\
        			\hline
        			V102 & 503.26 kB/s & 2.33 kB/s & 570.66 ms \\
        			\hline
        			V103 & 609.22 kB/s & 2.31 kB/s & 850.24 ms \\
        			\hline
        		\end{tabular}
        	}
        \end{table}
\bibliographystyle{abbrv-doi}
\bibliography{main}